\documentclass[a4paper]{llncs} 

\usepackage{amssymb}
\usepackage{graphicx}

\usepackage{url}
\urldef{\mailsa}\path|cwl@aiai.edu.cn|
\newcommand{\keywords}[1]{\par\addvspace\baselineskip
\noindent\keywordname\enspace\ignorespaces#1}

\begin{document}


\title{Multi-granular Perspectives on Covering}


%
%
\author{Wan-Li Chen}
%

\institute{Department of Computer Engineering\\
Anhui Institute of Architecture and Industry,\\
230601 HeFei, Anhui Province, China\\
\mailsa\\
}

%
%

\maketitle

\begin{abstract}
Covering model provides a general framework for granular computing
in that overlapping among granules are almost indispensable.
For any given covering, both intersection and union of covering blocks containing an element
are exploited as granules to form granular worlds at different abstraction levels, respectively,
and transformations among these different granular worlds are also discussed.
As an application of the presented multi-granular perspective on covering,
relational interpretation and axiomization of four types of covering based rough upper approximation operators
are investigated, which can be dually applied to lower ones.
\end{abstract}

\section{Introduction}

The ability to conceptualize the world at different granularities and to switch among these granularities
is fundamental to human intelligence and flexibility \cite{Hobbs}\cite{ZhangL}.
To simulate such an ability of human problem solving, Granular Computing (GrC) was proposed.
In fact, there have existed many concrete models of granular computing,
such as computing with words \cite{Zadeh96}\cite{Zadeh97}, rough set theory \cite{Pawlak},
formal concept analysis \cite{Ganter}, quotient space theory \cite{ZhangL}, etc.

Granulation \cite{Zadeh97}\cite{Lin}, grouping some elements of the domain into granules,
is a fundamental step towards granular computing.
Idealized granulation should yield pairwise disjoint granules such as equivalence classes or partition blocks.
However, incompleteness and impreciseness of information, as well as variant similarity of individuals,
always result in overlapping among granules.
Consequently, granular worlds are always formalized as
a set equipped with a family of nonempty subsets whose union equals to the universe,
which is precisely the covering model \cite{Lin}.

For any given covering, both union and intersection of covering blocks containing the given element
can be viewed as induced granules, which form two new coverings
representing granular worlds at different abstraction levels.
Consequently, transformations among these three covering are important and interesting
in that they simulate the process of switching among granularities,
which are precisely our respect of discussion.

Pawlak's roughs set provides a concrete framework performing granular computing based on partition or equivalence relation,
however, absolute disjointness among granules restrict applications of classical rough set theory,
so generalized rough set models based on covering are extensively researched,
and many results such as axiomization and reduction theory are obtained
(see, for example, in \cite{ZhangY}\cite{Zhu03}\cite{Zhu07}\cite{Zhu09} for details).
In these researches, union and intersection of covering blocks containing the given element
were always called the friend and neighborhood, respectively,
and they were exploited as basic knowledge to approximate unknown knowledge.
However, transformations among induced granular worlds as well as relational interpretations of friend and neighborhood
are not discussed, while the latter naturally results in
axiomization of several types of covering based rough approximation operators.

The paper is organized as follows:
Based on the definition of star and closure of an element,
two granular worlds are induced in section 2, and transformations among them are also discussed.
Section 3 investigates axiomization and relational interpretation
of four types of covering rough approximation operators,
and it is proved that these operators exploit induced granules, rather than given covering blocks,
as basic knowledge to approximate unknown knowledge.
Conclusions and discussions are grouped in the final section.

Throughout this paper, we do not restrict the universe of discourse $U$ to be finite.
The class of all subsets of $U$ will be denoted by $\wp(U)$.
For any $X \subseteq U$, $-X$ will be used to denote the complement of $X$.
For any subset system $ \mathcal{F} \subseteq \wp(U)$,
we will denote $ \{ - F | \: F \in \mathcal{F} \}$ as $\mathcal{F}^c$.
Obviously, $(\mathcal{F}^c)^c = \mathcal{F}$.
Moreover, $I$ always denotes arbitrary index set.

\section{Granular Worlds Induced by Covering}

In this section, we firstly recall the star and point closure system of any given covering,
and then we discuss transformations among these three subset systems.
When coverings are understood as granular worlds at some abstraction levels,
the transformations are precisely the processes of switching among different granularities.

\subsection{Star and Point Closure System of Covering}
A family $\beta$ of nonempty subsets of $U$ is called a covering of $U$
if $ \bigcup \{ B | \: B \in \beta \} = U $,
and any $ B \in \beta $ is called a covering block.
Obviously, covering generalizes partition in that overlapping among covering blocks are permitted.
For any $ x \in U$, let $\beta_x$ denote the collection of covering blocks containing $x$,
namely $\beta_x = \{ B | \: x \in B \in \beta \}$.

\begin{definition}[\cite{Cech}\cite{Erne}]
\label{def-StarPointClosure}
For any $ x \in U$, the star and point closure of $x$ with respect to $\beta$,
denoted by $ star(x,\beta) $ and $ \downarrow \hspace*{-1mm} x $,
are defined as the union and intersection of covering blocks containing $x$, respectively.
\end{definition}

Formally, $ \forall x \in U$, $ star(x,\beta) = \bigcup \{ B | \: B \in \beta_x \} $ and
$ \downarrow \hspace*{-1mm} x = \bigcap \{ B | \: B \in \beta_x \} $.
In \cite{Zhu07}\cite{ZhangY},
$star(x,\beta)$ and $\downarrow \hspace*{-1mm} x$ are called friend and neighborhood of $x$,
and denoted by $Friends(x)$ and $Neighbor(x)$, respectively.
In the followings, the collection of stars and point closures with respect to covering $\beta$
are denoted by $\mathcal{S}(\beta)$ and $\mathcal{P}(\beta)$,
namely $ \mathcal{S}(\beta) = \{ star(x,\beta) | \: x \in U \} $ and
$ \mathcal{P}(\beta) = \{ \downarrow \hspace*{-1mm} x | \: x \in U \} $.
Both the star system and the point closure system are coverings of $U$
since $ x \in star(x,\beta)$ and $ x \in \downarrow  \hspace*{-1mm} x $ hold for any $x \in U$.

\begin{definition}[\cite{Cech}]
\label{def-refinement}
For any coverings $ \alpha $ and $ \beta $, if for any $ A \in \alpha $ there exists $ B \in \beta $
such that $ A \subseteq B $, then $ \alpha $ is called a refinement of $ \beta $, denoted by $ \beta \sqsubseteq \alpha$.
\end{definition}

\begin{proposition}
\label{prop-refinement}
$ \mathcal{S}(\beta) \sqsubseteq \beta \sqsubseteq \mathcal{P}(\beta) $.
\end{proposition}

For any $ x \in U $, $star(x,\beta)$ and $\downarrow \hspace*{-1mm} x$ are regarded as granules of different grain size,
then Proposition \ref{prop-refinement} implies that
the two induced coverings represent granular worlds at different abstraction levels.

\begin{definition}[\cite{Erne}]
\label{def-SpecializationPreorder}
For any covering $ \beta $, the specialization preordering $ \preceq_\beta $ induced by $\beta$ is defined as:
$ \forall x,y \in U $, $ y \preceq_\beta x \Leftrightarrow (\forall K \in \beta)(x \in K \rightarrow y \in K ) $.
\end{definition}

In general, specialization preordering $ \preceq_\beta $ is reflexive and transitive but not necessarily antisymmetric.
Moreover, point closure $ \downarrow \hspace*{-1mm} x $ is precisely the principal down-set $ \{ y | \: y \preceq_\beta x \} $
generated by $x$ with respect to $ \preceq_\beta $.
The following Proposition \ref{prop-PointStructure} characterizes the structure of $\mathcal{P}(\beta)$.

\begin{proposition}
\label{prop-PointStructure}
For any covering $ \beta $ of $U$, $ \forall x,y \in U $, then
\begin{enumerate}
   \item $ y \in \downarrow \hspace*{-1mm} x $ if and only if  $ \downarrow \hspace*{-1mm} y \subseteq \downarrow \hspace*{-1mm} x$.
   \item $ \downarrow \hspace*{-1mm} x = \bigcup \{ \downarrow \hspace*{-1mm} z | \: z \in \downarrow \hspace*{-1mm} x \} $,
         $ \downarrow \hspace*{-1mm} x = \bigcap \{ \downarrow \hspace*{-1mm} z | \: x \in \downarrow \hspace*{-1mm} z \} $.
\end{enumerate}
\end{proposition}

For further discussion, we need to recall some notations about binary relation.
For any binary relation $R$ on $U$ and $x \in U$,
$R(x)= \{ y \in U | \: xRy\}$ and $R^{-1} = \{ \langle v,u \rangle | \: uRv \}$.
Particularly, when $\leq$ is a partial ordering or preordering,
$\leq^{-1}$ is always called the dual of $\leq$ and denoted by $\geq$.
Moreover, $\uparrow \hspace*{-1mm} x = \{ y \in U | \: x \leq y\}$
and $\downarrow \hspace*{-1mm} x = \{ y \in U | \: y \leq x \}$
are called principal up-set and principal down-set generated by $x$, respectively \cite{Erne}.

For any covering $ \beta $, the subset system $ \beta^c = \{ -B | \: B \in \beta \}$
is not necessarily a covering of $U$ any more.
In fact, $ \beta^c $ is also a covering of $U$ if and only if $ \bigcap \{ B | \: B \in \beta\} = \emptyset$,
or equivalently, there exist $B \in \beta$ such that $x \notin B$ holds for any $x \in U$.
However, the specialization preordering $\preceq_{\beta^c}$ can be still defined,
and we have the following proposition characterizing the relationship between $\preceq_\beta$ and $\preceq_{\beta^c}$.

\begin{proposition}[\cite{Erne}]
\label{prop-ComPointStructure}
For any covering $ \beta $ of $U$, the specialization preordering $\preceq_{\beta^c}$
is precisely the dual of $\preceq_\beta$, namely $ \preceq_{\beta^c} = \succeq_\beta $.
\end{proposition}

\begin{definition}[\cite{Erne}]
\label{def-Core}
For any covering $ \beta $ of $U$, $x \in U$,
the point closure of $x$ with respect to the subset system $\beta^c$ is called the core of $x$,
denoted by $ \uparrow \hspace*{-1mm} x $.
\end{definition}

For any $x \in U$, $ \uparrow \hspace*{-1mm} x = \bigcap \{ D | \: x \in D \in \beta^c \}$
and $\mathcal{P}(\beta^c) = \{ \uparrow \hspace*{-1mm} x | \: x \in U \}$.
It is easy to verify that $\uparrow \hspace*{-1mm} x $ is precisely the principal up-set
$ \{ y \in U \ | \: x \preceq_\beta y \}$ generated by $x$ with respect to $\preceq_\beta$.
Note that $\mathcal{P}(\beta^c)$ is also a covering of $U$
since $x \in \uparrow \hspace*{-1mm} x$ holds for any $x \in U$.

Any covering $ \beta $ can also induces a binary relation $ T_\beta $
determined by $ \bigcup \{ B \times B | \: B \in \beta \} $.
Moreover, $ \forall x \in U$, $ T_\beta (x) = star(x, \beta) $.
Then we have the following relational characterizations of the three induced coverings.

\begin{theorem}
\label{thm-RelationalCharacterization}
For any covering $\beta$, point closure systems
$ \mathcal{P}(\beta) = \{ \succeq_\beta \hspace*{-1mm} (x) | \: x \in U \} $,
$ \mathcal{P}(\beta^c) = \{ \preceq_\beta \hspace*{-1mm} (x) | \: x \in U \} $,
and star system $ \mathcal{S}(\beta) = \{ T_\beta(x) | \: x \in U \} $.
\end{theorem}

\subsection{Transformations among Induced Granular Worlds}

Note that, according to Proposition \ref{prop-refinement},
$\mathcal{P}$ transforms a coarser granular world into finer one,
while $\mathcal{S}$ transforms a finer granular world into coarser one,
so they can be understood as unary operations on covering
simulating the process of switching among different granular worlds.
Proposition \ref{prop-PointStructure} only tells us that
$\mathcal{P}(\mathcal{P}(\beta)) = \mathcal{P}(\beta)$,
namely $\mathcal{P}$ is an idempotent operation.
To completely understand these transformations,
we need to study $\mathcal{P}$ and $\mathcal{S}$ further.

For any covering $\beta$, $T_\beta$ is a tolerance,
a binary relation satisfying reflexivity and symmetry \cite{Bartol}.
Theorem \ref{thm-RelationalCharacterization} implies that
tolerance relation will be an useful tool to investigate properties of $\mathcal{S}$,
so we will firstly recall some basic definitions and facts about tolerance relation.

Let $T$ be any tolerance relation on $U$, $x \in U$, the $T$-relative set $T(x)$ is always called $T$-class,
and we will denote the family of all $T$-classes as $\mathfrak{C}(T)$, namely $\mathfrak{C}(T) = \{ T(x) | \: x \in U\}$.
Obviously, $\mathfrak{C}(T)$ is an covering induced by $T$,
and for any covering $\beta$, $ \mathcal{S}(\beta) = \mathfrak{C}(T_\beta)$.

A $T$-preblock is any subset $B \subseteq U$ such that $xTy$ holds for any $ x,y \in B $.
Any subset $B \subseteq U$ is called a $T$-block if and only if it is a $T$-preblock such that
for any $ x \notin B $, there exists $b \in B$ satisfying $ \neg (bTx) $.
In the current paper, the family of all $T$-blocks will be denoted as $\mathfrak{B}(T)$,
which is also an induced covering by $T$.
It is well known that there exists one-to-one correspondence between
the collection of all tolerance relations and the collection of all $T$-blocks on $U$.
Moreover, we have the following Proposition \ref{prop-BlockClass},
which not only characterizes the relationship between $T$-classes and $T$-blocks,
but also implies that $\mathfrak{C}(T) = \mathcal{S}(\mathfrak{B}(T))$.

\begin{proposition}[\cite{Bartol}]
\label{prop-BlockClass}
Let $(U,T)$ be any tolerance space, $ x \in U$, $ B \in \mathfrak{B}(T)$, then
$ T(x) = \bigcup \{B | \: x \in B \} $, $ B = \bigcap \{T(x) | \: \ x \in B \} $.
\end{proposition}

For any tolerance space $(U,T)$ and $ x \in U $, the intersection of $T$-classes containing $x$ was called
compatibility kernel $\langle x \rangle_T$ in \cite{Cheng},
namely $ \langle x \rangle _T = \bigcap \{T(y) | \: x \in T(y)\}$,
and the family of all compatibility kernels was denoted by $\langle U \rangle_T$.
Obviously, $\langle x \rangle_T = \downarrow \hspace*{-1mm} x$,
in which $\downarrow \hspace*{-1mm} x $ is the point closure of $x$ with respect to covering $\mathfrak{C}(T)$.
Furthermore, we have $\langle U \rangle_T = \mathcal{P}(\mathfrak{C}(T))$.
It was also proved in \cite{Cheng} that $\langle x \rangle_T = \bigcap \{B | \: x \in B \in \mathfrak{B}(T)\}$,
or equivalently, $\mathcal{P}(\mathfrak{B}(T))=\mathcal{P}(\mathfrak{C}(T))$.
So in the following, the point closure system $\mathcal{P}(T)$ of a tolerance relation $T$
is always referred to $\mathcal{P}(\mathfrak{B}(T))$ or $\mathcal{P}(\mathfrak{C}(T))$.

\begin{proposition}
\label{prop-PBC}
Let $(U,T)$ be any tolerance space, $ x,y \in U$, $ B \in \mathfrak{B}(T)$, then
\begin{enumerate}
   \item $ B = \bigcup \{ \downarrow \hspace*{-1mm} y | \: y \in B \} $,
         $ T(x) = \bigcup \{ \downarrow \hspace*{-1mm} y  | \: y \in T(x) \} $.
   \item $ \bigcup \{ \downarrow \hspace*{-1mm} y | \: x \in \downarrow \hspace*{-1mm} y \} \subseteq T(x) $.
\end{enumerate}
\end{proposition}

The proofs of the above proposition is trivial, however, it should be pointed out that
$ \bigcup \{ \downarrow \hspace*{-1mm} y | \: x \in \downarrow \hspace*{-1mm} y \} \supseteq T(x) $
does not hold in general.
Now we turn to discuss the relationship between
$\mathcal{P}(\mathcal{S}(\beta))$ and $\mathcal{P}(\beta)$.

\begin{proposition}
\label{prop-PSBeta}
For any covering $ \beta $ of $U$,
$ \mathcal{P}(\mathcal{S}(\beta)) = \mathcal{P}(T_\beta) \sqsubseteq \mathcal{P}(\beta) $.
Furthermore, if $ \beta $ is precisely $\mathfrak{B}(T)$ with respect to tolerance $T$,
then $ \mathcal{P}(\mathcal{S}(\beta)) = \mathcal{P}(\beta) $.
\end{proposition}

\begin{proof}
It is sufficient to prove that
$ \downarrow \hspace*{-1mm} x \subseteq \langle x \rangle _{T_\beta} $ holds for any $x \in U$.
Note that $ y \in \downarrow \hspace*{-1mm} x \Leftrightarrow \forall B \in \beta (x \in B \rightarrow y \in B)$,
then if $ x \in T_\beta (z) = \bigcup \{ B | \: z \in B \in \beta \} $,
there must exists $B_0 \in \beta$ such that $x \in B_0$ and $z \in B_0$.
So $y \in B_0$ and $z \in B_0$ holds, which follows that $y \in T_\beta (z)$.
Hence, $\forall z \in U (x \in T_\beta(z) \rightarrow y \in T_\beta(z))$,
namely $ \downarrow \hspace*{-1mm} x \subseteq \langle x \rangle _{T_\beta} $.

If $ \beta $ is precisely $\mathfrak{B}(T)$ with respect to tolerance relation $T$,
then $ \downarrow \hspace*{-1mm} x = \langle x \rangle _{T_\beta} $ holds for any $x \in U$.
\end{proof}

\begin{remark}
\textup{The converse of proposition \ref{prop-PSBeta} is not true in general.
Let $ U = \{1,2,3\}$ and $ \beta = \{ \{1,3\},\{2,3\},\{3\} \}$,
then $ \downarrow \hspace*{-1mm} x = \langle x \rangle $ holds for any $ x \in U $,
however, $ \beta $ is not the family of tolerance blocks at all.}
\end{remark}

For any tolerance relation $T$, $ \mathfrak{C}(T) $, $ \mathfrak{B}(T) $ and $ \mathcal{P}(T) $
are all coverings of $U$ induced by $T$, which represent granular worlds at different abstraction level.
We also have known that $ \mathcal{P}(\mathfrak{C}(T)) = \mathcal{P}(T) = \mathcal{P}(\mathfrak{B}(T)) $
and $ \mathcal{S}(\mathfrak{B}(T)) = \mathfrak{C}(T) $.
In general, however, $ T_{\mathcal{P}(T)}(x) \subseteq T(x)$ for any $ x \in U $,
or equivalently $ \mathfrak{C}(T) \sqsubseteq \mathcal{S}(\mathcal{P}(T)) $,
in which $ T_{\mathcal{P}(T)} $ denotes the tolerance relation
induced by $ \mathcal{P}(T) = \{ \langle x \rangle _T | x \in U \} $.
The following theorem characterizes sufficient and necessary condition of
$ \mathcal{S}(\mathcal{P}(T)) = \mathfrak{C}(T) $, or equivalently $ T= T_{\mathcal{P}(T)} $.

\begin{theorem} [\cite{Cheng}]
\label{thm-Cheng}
For any tolerance relation $T$, then $ T_{\mathcal{P}(T)} = T $ if and only if,
for any $a,b \in U $, $aTb$ implies that there exists $c \in U$ such that
$ a \in \langle c \rangle_T $ and $ b \in \langle c \rangle_T $.
\end{theorem}

An interesting conclusion of Proposition \ref{prop-PSBeta} and Theorem \ref{thm-Cheng} is
$ \mathcal{S}(\mathcal{P}(\mathcal{S}(\mathcal{P}(\alpha)))) = \mathcal{S}(\mathcal{P}(\alpha)) $
holds for any covering $\alpha$,
which implies that the composition of $\mathcal{P}$ and $\mathcal{S}$ is also idempotent.

\begin{theorem}
\label{thm-SPSBeta}
For any covering $ \alpha $ of $U$, we denote $\mathcal{P}(\alpha) $ by $\beta$,
then $ T_{\mathcal{P}(T_\beta)} = T_\beta$.
\end{theorem}

\begin{proof}
Note that $ \mathcal{P}(\mathcal{P}(\alpha)) = \mathcal{P}(\alpha) $ and $ \beta = \mathcal{P}(\alpha) $,
then $\downarrow_\alpha \hspace*{-1mm} x = \downarrow_\beta \hspace*{-1mm} x \triangleq \downarrow \hspace*{-1mm} x $,
in which $\downarrow_\alpha \hspace*{-1mm} x$ and $\downarrow_\beta \hspace*{-1mm} x$
denote the point closure of $x$ with respect to $\alpha$ and $\beta$, respectively.
By the proof of Proposition \ref{prop-PSBeta},
$\downarrow \hspace*{-1mm} x  \subseteq \langle x \rangle _{T_\beta} $ holds for any $x \in U$.

For any $a,b \in U $, since $ T_\beta(x) = \bigcup \{ \downarrow \hspace*{-1mm} y | \: x \in \downarrow \hspace*{-1mm} y \}$,
if $ a T_\beta b $ then there exists $c \in U $ such that
$a \in \downarrow \hspace*{-1mm} c $ and $b \in \downarrow \hspace*{-1mm} c $,
so $a \in \langle c \rangle _{T_\beta} $ and $b \in \langle c \rangle _{T_\beta} $.
According to Theorem \ref{thm-Cheng}, we have $ T_{\mathcal{P}(T_\beta)} = T_\beta$.
\end{proof}

\section{Applications to Covering Rough Approximation}

This section mainly discusses some applications of our multi-granular perspectives on covering.
we firstly recall relationship between quasi-discrete closure operator and binary relation,
and then discuss relational interpretations as well as axiomizations
of four types of covering rough approximation operators.

\subsection {C\v{e}ch Closure Operators}
Closure operators in the sense of C\v{e}ch \cite{Cech}\cite{Galton}
generalize topological ones in that idempotent axiom does not necessarily holds.

\begin{definition}[\cite{Cech}\cite{Galton}]
\label{def-CechClosure}
Let $Cl : \wp(U) \rightarrow \wp(U) $ be any mapping, $ \forall X,Y \subseteq U $,
$Cl$ is called C\v{e}ch closure operator on $U$ if it satisfies following axioms (C1)-(C3): \\
\hspace*{0.5in} (C1) \hspace{0.01in} $ Cl(\emptyset) = \emptyset $              \\
\hspace*{0.5in} (C2) \hspace{0.01in} $ X \subseteq Cl(X)$                       \\
\hspace*{0.5in} (C3) \hspace{0.01in} $ Cl(X \bigcup Y) = Cl(X) \bigcup Cl(Y) $  \\
If, in addition, $Cl$ also satisfies quasi-discreteness axiom (C4),
then we call it C\v{e}ch quasi-discrete closure operator: \\
\hspace*{0.5in} (C4) \hspace{0.01in} $ Cl(X) = \bigcup \{ Cl(x)| \: x \in X \} $\\
in which $Cl(x)$ denotes the closure of singleton $\{x\}$.
\end{definition}

Note that (C4) is eqivalent to the following axiom (C4'): \\
\hspace*{0.5in} (C4') \hspace{0.01in} $ Cl(\bigcup\{X_i | \: i \in I \}) = \bigcup \{ Cl(X_i)| \: i \in I \} $ \\
where $ X_i \subseteq U$ for any $i \in I$.
Any C\v{e}ch closure operator $Cl$ satisfying Kurotowski axiom (C5): \\
\hspace*{0.5in} (C5) \hspace{0.01in} $ Cl(X) = Cl(Cl(X)) $ \\
is called topological closure operator,
and any quasi-discrete topological closure operator is called Alexzandroff closure operator.

Let $Cl$ be any C\v{e}ch closure operator on $U$, $ \forall X \subseteq U$,
the inetrior of $X$ is defined as the complement of the closure of the complement of $X$.
Formally, $Int(X) = -(Cl(-X))$, and we always call $Int$  the interior operator.
For any $x \in U$, if $x \in Int(X)$ then $X$ is called a neighbourhood of $x$.

\begin{proposition}
\label{prop-ClIntNeigh}
Let $Cl$ be any C\v{e}ch closure operator on $U$, $X,X_i \subseteq U$ and $x \in U$,
\begin{enumerate}
   \item $ Cl(X) = \bigcup \{ Cl(x)| \: x \in U \} $ if and only if each element $u$ of $U$ has a minimal neighbourhood $N(u)$.
   \item $ Cl(\bigcup\{X_i | \: i \in I \}) = \bigcup \{ Cl(X_i)| \: i \in I \} $ if and only if
         $ Int(\bigcap\{X_i | \: i \in I \}) = \bigcap \{ Int(X_i)| \: i \in I \} $.
\end{enumerate}
\end{proposition}

Moreover, C\v{e}ch quasi-discrete closure operators are closely connected with binary relations on $U$.

\begin{theorem}[\cite{Galton}]
\label{thm-CloRel}
Let $ Cl: \wp(U) \rightarrow \wp(U) $, then $Cl$ is C\v{e}ch quasi-discrete closure operator if and only if
there exists binary relation $ R \subseteq U \times U $ such that, for any $ X \subseteq U $,
$ Cl(X)= Cl_R(X)= X \bigcup \{x \in U | \: R(x) \bigcap X \neq \emptyset \}$.
\end{theorem}

For any relation $R$ on $U$, let $Int_R$ denote the dual of $Cl_R$,
then $Int_R(X) =-Cl_R(-X)=\{ x \in X | \: R(x) \subseteq X \}$,
$N(u)=\{u\} \bigcup R(u)$,
$Cl_R(u) = \{u\} \bigcup R^{-1}(u)$.
By Theorem \ref{thm-CloRel}, if $R$ is reflexive, $Cl_R(X) = \{x \in U | \: R(x) \bigcap X \neq \emptyset \}$.
In fact, any reflexive relation on $U$ bijectively corresponds to C\v{e}ch quasi-discrete closure operator \cite{Cech}\cite{Galton},
which generates the well-known 1-1 correspondence between preordering relations and Alexandroff topologies.

\begin{proposition}
\label{prop-SymRel}
For any relation $R$ on $U$, $R$ is symmetric if and only if for any $u \in U$,
$Cl_R(u)$ is the minimal neighbourhood $N(u)$ of $u$.
\end{proposition}

\subsection{Covering based Rough Approximation Operators}
We first recall a type of generalized rough approximation operators based on arbitrary binary relation.

\begin{definition}[\cite{Yao}]
\label{def-Rel}
Let $R$ be any binary relation on $U$, operators $\overline{R}, \underline{R} : \wp(U) \rightarrow \wp(U) $
are defined as follows: $ \forall X \subseteq U $, \\
\hspace*{0.5in} $ \overline{R}(X)=  \{ x \in U | \: R(x) \bigcap X \neq \emptyset \}$ \\
\hspace*{0.5in} $ \underline{R}(X)= \{ x \in U | \: R(x) \subseteq X \} $ \\
We call $\overline{R}, \underline{R}$ generalized rough upper and lower approximation operators based on relation $R$.
\end{definition}

Obviously, for any binary relation $R$ on $U$, $X \subseteq U$,
$Cl_R(X) = X \bigcup \overline{R}(X)$ (or equivalently, $ Int_R(X) = X \bigcap \underline{R}(X) $).
Particularly, if $R$ is reflexive then $Cl_R(X) =\overline{R}(X)$
(or equivalently, $ Int_R(X) = \underline{R}(X) $).

For any given covering space, different types of covering based rough approximation operators
have been proposed \cite{Bonikowski}\cite{Qin}\cite{Zhu03}\cite{Zhu09}.
In this subsection, we will mainly discuss relational interpretations and axiomatic characterizations
of four types of covering based rough approximation operators.

\begin{definition}[\cite{Zhu07}\cite{ZhangY}]
\label{def-First}
For any covering $\beta $ of $U$, operators $FH, FL : \wp(U) \rightarrow \wp(U) $
are defined as follows: $ \forall X \subseteq U $, \\
\hspace*{0.5in} $ FH(X)= \bigcup\{ B \in \beta | \: B \bigcap X \neq \emptyset \}$ \\
\hspace*{0.5in} $ FL(X)\: = \{ x \in U | \: \forall B \in \beta (x \in B \rightarrow B \subseteq X) \} $ \\
We call $FH$, $FL$ the first type of covering upper and lower operators, respectively.
\end{definition}

It is obvious that $FH$ and $FL$ are dual.
Moreover, we have following relational interpretation of $FH$ and $FL$.

\begin{proposition}
\label{prop-First1}
For any covering $\beta $ of $U$, let $T_\beta$ denote the induced tolerance relation,
then $ \forall X \subseteq U $, \\
\hspace*{0.5in} $ FH(X)= \{ x \in U | \: T_{\beta}(x) \bigcap X \neq \emptyset \} $ \\
\hspace*{0.5in} $ FL(X)\: = \{ x \in U | \: T_{\beta}(x) \subseteq X \} $
\end{proposition}

\begin{proof}
Since $FH$ and $FL$ are dual operators, it is sufficient to prove only the half.
$ \forall y \in \bigcup\{ B \in \beta | \: B \bigcap X \neq \emptyset \}$,
there exists $ B_0 \in \beta $ and $y \in B_0$,
which follows $ T_{\beta}(y) \bigcap X \neq \emptyset $,
then $ y \in \{ x \in U | \: T_{\beta}(x) \bigcap X \neq \emptyset \} $.
On the other hand, $\forall y \in \{ x \in U | \: T_{\beta}(x) \bigcap X \neq \emptyset \}$,
there exists $B_0 \in \beta$ such that $y \in B_0$ and $ B_0 \bigcap X \neq \emptyset$,
then $ y \in \bigcup\{ B \in \beta | \: B \bigcap X \neq \emptyset \}$.
\end{proof}

\begin{proposition}
\label{prop-First2}
Let $T_\beta$ be the induced tolerance relation of covering $\beta$, $\forall X \subseteq U$,
then $ FH(X)= Cl_\beta(X)$ and $FL(X)= Int_\beta(X)$,
in which $Cl_ \beta$ and $Int_ \beta$ are C\v{e}ch quasi-discrete closure and interior operators
bijectively corresponding to tolerance relation $T_\beta$, respectively.
\end{proposition}

Proposition \ref{prop-First2} implies that $FH$ and $FL$ are generalized rough upper and lower approximation operators
based on tolerance relation $T_\beta$, or equivalently,
granules $T_\beta(x) = star(x,\beta)$ of induced granular world $\mathcal{S}(\beta)$
are exploited as basic knowledge blocks to approximate uncertain knowledge.
Furthermore, we have the following axiomatic characterization of $FH$.

\begin{theorem}
\label{thm-First}
For any mapping $ H: \wp(U) \rightarrow \wp(U) $, $\forall x,y \in U$, $X \subseteq U$ and
$ X_i \subseteq U \ (\forall i \in I)$,
then $H$ satisfies the following properties (1H)-(4H): \\
\hspace*{0.5in} (1H) \hspace{0.01in} $ H(\emptyset) = \emptyset $              \\
\hspace*{0.5in} (2H) \hspace{0.01in} $ X \subseteq H(X)$                       \\
\hspace*{0.5in} (3H) \hspace{0.01in} $ H(\bigcup X_i) = \bigcup H(X_i) $        \\
\hspace*{0.5in} (4H) \hspace{0.01in} $ y \in H(x) \Leftrightarrow x \in H(y) $  \\
if and only if there exists a covering $\beta$ such that $ H = FH $.
\end{theorem}

\begin{proof}
Suppose $H$ satisfies properties (1H)-(4H).
We define $T \subseteq U \times U$:
$ \forall x,y \in U $, $xTy \Leftrightarrow y \in H(x)$,
then $T$ is tolerance relation and $T(x)=H(x)$ holds for any $ x \in U$.
Let $\beta= \mathfrak{B}(T)$, then $ FH = Cl_{T_\beta}$ and
$ FH(x)=\bigcup \{ B \in \beta | \: x \in B \} =T(x) =H(x)$.
Furthermore, considering the quasi-discreteness of $Cl_{T_\beta}$,
$\forall X \subseteq U$, we have
$ FH(X) =\bigcup FH(x) =\bigcup H(x) =H(X) $,
which follows $FH=H$.

By Proposition \ref{prop-First2}, $FH(X) = Cl_\beta(X)$ holds for any $X \subseteq U$,
so the proof of the sufficiency is trivial.
\end{proof}

\begin{remark}
\textup{In the proof of sufficiency, the covering $\beta$ is not unique.
For example, the family of binary subsets $ \{x,y\} $ such that
$y \in H(x)$ is chosen as the covering $\beta$ in \cite{ZhangY}.}
\end{remark}

The second type of covering rough approximation operators
can be regarded as a special case of the first ones
in the sense that the covering is an induced point closure system,
however, their axiomatic characterization are different.

\begin{definition}[\cite{Zhu03}\cite{ZhangY}]
\label{def-Second}
For any covering $ \beta $, operators $SH, SL : \wp(U) \rightarrow \wp(U) $
are defined as follows: $ \forall X \subseteq U $, \\
\hspace*{0.5in} $ SH(X)= \bigcup\{ \downarrow \hspace*{-1mm} x | \: \downarrow \hspace*{-1mm} x \bigcap X \neq \emptyset \}$ \\
\hspace*{0.5in} $ SL(X)\: = \{ x \in U | \: \forall y \in U (x \in \downarrow \hspace*{-1mm} y \rightarrow \downarrow \hspace*{-1mm} y \subseteq X) \}$ \\
We call $SH$, $SL$ the second type of covering upper and lower operators, respectively.
\end{definition}

Then, in a completely similar way, we have the following Proposition \ref{prop-Second1} and Proposition \ref{prop-Second2}.

\begin{proposition}
\label{prop-Second1}
For any covering $\beta $ of $U$, let $T_{\mathcal{P}(\beta)}$ denote the induced tolerance relation,
then $ \forall X \subseteq U $, \\
\hspace*{0.5in} $ SH(X)= \{ x \in U | \: T_{\mathcal{P}(\beta)}(x) \bigcap X \neq \emptyset \} $ \\
\hspace*{0.5in} $ SL(X)\: = \{ x \in U | \: T_{\mathcal{P}(\beta)}(x) \subseteq X \} $
\end{proposition}

\begin{proposition}
\label{prop-Second2}
For any $ X \subseteq U$, $ SH(X)= Cl_{\mathcal{P}(\beta)}(X) $ and $ SL(X)= Int_{\mathcal{P}(\beta)}(X) $,
in which $Cl_{\mathcal{P}(\beta)}$ and $Int_{\mathcal{P}(\beta)}$ are C\v{e}ch quasi-discrete closure and interior operators
bijectively corresponding to $T_{\mathcal{P}(\beta)}$, respectively.
\end{proposition}

Proposition \ref{prop-Second2} implies that $SH$ and $SL$ are generalized rough upper/lower approximation operators
based on tolerance relation $T_{\mathcal{P}(\beta)}$, or equivalently,
granules $T_{\mathcal{P}(\beta)}(x)$ of induced granular world $\mathcal{S}(\mathcal{P}(\beta))$ are basic knowledge blocks.

\begin{theorem}
\label{thm-Second}
Let $ H: \wp(U) \rightarrow \wp(U) $, $\forall x,y \in U$, $X \subseteq U$ and
$ X_i \subseteq U $ holds for any $i $ of arbitrary index set $I$,
then there exists a covering $\beta$ such that $ H = SH $ if and only if
$H$ satisfies (1H)-(4H) and the following property (5H): \\
\hspace*{0.1in} (5H) \hspace{0.01in} if $ y \in H(x)$, there exists $u \in U$  such that $ x,y \in \bigcap \{H(z) | \: u \in H(z) \} $
\end{theorem}

\begin{proof}
Suppose $H$ satisfies properties (1H)-(4H).
We define $T \subseteq U \times U$:
$ \forall x,y \in U $, $xTy \Leftrightarrow y \in H(x)$,
then $T$ is tolerance relation and $T(x)=H(x)$ holds for any $ x \in U$.
Note that $\langle x \rangle_T = \bigcap \{H(z) | \: x \in H(z) \}$,
so according to Theorem \ref{thm-Cheng}, we have $T=T_{\mathcal{P}(T)}$.

Let $\beta= \mathfrak{B}(T)$, then $T_{\mathcal{P}(\beta)} =T_{\mathcal{P}(T)} =T$,
so for any $ x \in U$,
$ SH(x)=\bigcup \{ \downarrow \hspace*{-1mm} y| \: x \in \downarrow \hspace*{-1mm} y \} =T(x) =H(x)$.
Considering the quasi-discreteness of $Cl_{\mathcal{P}(\beta)}$,
we have $SH(X) =\bigcup \{SH(x) | \: x \in X\} =\bigcup \{H(x) | \: x \in X\} =H(X)$,
which follows $SH=H$.

Suppose $\beta$ be any covering of $U$, then $ SH = Cl_{\mathcal{P}(\beta)}$.
According to Theorem \ref{thm-First}, $SH$ satisfies axioms (1H)-(4H);
Moreover, by theorem \ref{thm-Cheng} and theorem \ref{thm-SPSBeta}, $SH$ also satisfies axiom (5H).
\end{proof}

\begin{remark}
\textup{It should be pointed out that the covering $\beta$ is also not unique in the proof of sufficient condition.
For example, the family of subsets $ \{ H(x) | x \in U\} $ is chosen as the covering $\beta$ in \cite{ZhangY}.}
\end{remark}

The third type of covering based rough approximation operators on finite universe was defined in \cite{Zhu07},
however, the upper and the lower approximation operator were not dual.
In the following, we will discuss the third type of covering based upper approximation operator and its dual,
and our results also apply to infinite universe \cite{Liu}.

\begin{definition}[\cite{Zhu07}]
\label{def-third}
For any covering $\beta $ of $U$, operators $TH, TL : P(U) \rightarrow P(U) $
are defined as follows: $ \forall X \subseteq U $, \\
\hspace*{0.5in} $ TH(X)= \bigcup \{\: \downarrow \hspace*{-1mm} x | \: x \in X \} $ \\
\hspace*{0.5in} $ TL(X)\: = \{ x \in U | \: \forall u ( x \in \downarrow \hspace*{-1mm} u \rightarrow
                                                        \downarrow \hspace*{-1mm} u \subseteq X)\} $ \\
We call $TH$, $TL$ the third type of covering upper and lower operators, respectively.
\end{definition}

\begin{proposition}
\label{prop-third1}
For any covering $\beta $ of $U$, $ \forall X \subseteq U $,\\
\hspace*{0.5in} $ TH(X)= \{ x \in U | \uparrow \hspace*{-1mm} x \bigcap X  \neq \emptyset \} $ \\
\hspace*{0.5in} $ TL(X)\:= \{ x \in U | \uparrow \hspace*{-1mm} x \subseteq X \} $
\end{proposition}

\begin{proof}
For any $ X \subseteq U $, then
$TH(X)= \bigcup \{\: \downarrow \hspace*{-1mm} y | \: y \in X \} $
$     = \{ x \in U | \: \exists y \in X ( x \in \downarrow \hspace*{-1mm} y )\} $
$     = \{ x \in U | \: \uparrow \hspace*{-1mm}x \bigcap X \neq \emptyset \} $.
Dually, $ TL(X)= \{ x \in U | \uparrow \hspace*{-1mm} x \subseteq X \} $.
\end{proof}

According to Theorem \ref{thm-RelationalCharacterization},
$\uparrow \hspace*{-1mm} x = \preceq_\beta \hspace*{-1mm} (x)$,
then operators $TH$ and $TL$ are in fact C\v{e}ch quasi-discrete closure and interior operators
bijectively corresponding to the reflexive and transitive relation $\preceq_\beta$.

\begin{proposition}
\label{prop-third2}
For any $ X \subseteq U$, $TH(X)= Cl_{\preceq_\beta}(X)$, $TL(X) = Int_{\preceq_\beta}(X)$.
\end{proposition}

\begin{theorem}
\label{thm-Third}
Let $ H: P(U) \rightarrow P(U) $, $X \subseteq U$, $ X_i \subseteq U (\forall i \in I)$,
then $H$ satisfies the following properties (H1)-(H4): \\
\hspace*{0.5in} (H1) \hspace{0.01in} $ H(\emptyset) = \emptyset $              \\
\hspace*{0.5in} (H2) \hspace{0.01in} $ X \subseteq H(X)$                       \\
\hspace*{0.5in} (H3) \hspace{0.01in} $ H(\bigcup X_i) = \bigcup H(X_i) $        \\
\hspace*{0.5in} (H4) \hspace{0.01in} $ H(H(X)) = H(X) $  \\
if and only if there exists a covering $\beta$ such that $ H = TH $.
\end{theorem}

\begin{proof}
For any covering $\beta$, $TH=Cl_{\preceq_\beta}$, which follows that $TH$ satifies properties (H1)-(H4).

Suppose $H$ satisfies properties (H1)-(H4) and $ \forall x \in U$
By property (H2) then $x \in H(x)$ holds, which follows that $\beta = \{H(u) | \: u \in U \}$ is a covering of $U$.
By property (H4), $ H(H(x))=H(x) $ and $\downarrow \hspace*{-1mm} x = \bigcap \{ H(y) | \: x \in H(y) \}= H(x)$,
so $TH(x)=\{y | \: \uparrow \hspace*{-1mm} y \bigcap \{x\} \neq \emptyset\}
         = \{y | \: x \in \uparrow \hspace*{-1mm} y\} = \downarrow \hspace*{-1mm} x $.
For any $X \subseteq U$, by the quasi-discreteness of $TH$ as well as the property (H3),
then $TH(X) = \bigcup \{ TH(x) | \: x \in X \} = \bigcup \{ H(x) | \: x \in X \} = H(X) $.
\end{proof}

The fourth type of covering based rough approximation operators was defined in \cite{Xu}\cite{Zhu09},
and \cite{Chen}\cite{Wu} discussed a special case in that
the covering is precisely $\mathfrak{B}(T)$ of given tolerance relation $T$.
In \cite{Liu}\cite{Xu}, it was proved that the fourth type of operators are equivalent to
rough approximation operators based on preordering relation.

\begin{definition}[\cite{Zhu09}]
\label{def-Fourth}
For any covering $\beta $ of $U$, operators $XH, XL : P(U) \rightarrow P(U) $
are defined as follows: $ \forall X \subseteq U $,\\
\hspace*{0.5in} $ XH(X)= \{ x \in U | \: \downarrow \hspace*{-1mm} x \bigcap X \neq \emptyset \} $ \\
\hspace*{0.5in} $ XL(X)\: = \{ x \in U | \: \downarrow \hspace*{-1mm} x \subseteq X \} $ \\
We call $XH$, $XL$ the fourth type of covering upper and lower operators, respectively.
\end{definition}

According to Theorem \ref{thm-RelationalCharacterization},
$\downarrow \hspace*{-1mm} x = \succeq_\beta \hspace*{-1mm} (x)$,
so we have the following topological interpretation of $XH$ and $XL$.

\begin{proposition}
\label{prop-Fourth1}
For any $ X \subseteq U$, $XH(X)= Cl_{\succeq_\beta}(X)$, $XL(X) = Int_{\succeq_\beta}(X)$.
\end{proposition}

\begin{theorem}
\label{thm-Fourth}
Let $ H: P(U) \rightarrow P(U) $, $\forall x,y \in U$, $X \subseteq U$ and
$ X_i \subseteq U $ holds for any $i$ of arbitrary index set $I$,
then $H$ satisfies properties (H1)-(H4) if and only if there exists a covering $\beta$ such that $ H = XH $.
\end{theorem}

\begin{proof}
Suppose $H$ satisfies properties (H1)-(H4).
Let $\beta = \{K^c | H(K)=K \}$, then $\beta$ is a covering of $U$ since $ H(\emptyset) = \emptyset $,
and $\forall x \in U$, it is easy to verify that
the point closure of $x$ with respect to $\beta^c$ is precisely $H(x)$,
and that $XH(x)= \uparrow \hspace*{-1mm} x$, the core of $x$ with respect to $\beta$,
so $XH(x)= H(x)$.
For any $X \subseteq U$, by the quasi-discreteness of $XH$ and property (H3),
we have $XH(X) = \bigcup \{ XH(x) | \: x \in X \} = \bigcup \{ H(x) | \: x \in X \} = H(X) $.
\end{proof}

Note that the third and fourth types of upper approximation operators have the same axiomatic characterization,
however, the corresponding coverings are different.

\section{Conclusion}
Any covering can be associated with at least two new ones: point closure system and star system,
both of which have obvious relational interpretation.
The paper understands these coverings as granular worlds at different abstraction level,
and investigates typical properties of transformations among these granular worlds.

As an application of the presented multi-granular perspectives on covering,
four types of covering based rough approximation operators are proved to be induced by binary relations,
which, to some extent, bridges the gap between generalized rough sets based on binary relation and coverings.
Furthermore, axiomaic characterizations of these four types of upper approximation operators are obtained,
and these results can be dually apply to lower ones.

\end{document}